\documentclass{article}

\usepackage{PRIMEarxiv}

\usepackage[utf8]{inputenc} 
\usepackage[T1]{fontenc}    
\usepackage{hyperref}       
\usepackage{url}            
\usepackage{booktabs}       
\usepackage{amsfonts}       
\usepackage{nicefrac}       
\usepackage{microtype}      
\usepackage{lipsum}

\usepackage{adjustbox}
\usepackage{tabularx}
\newcommand{\inst}[1]{}
\newcommand{\institute}[1]{}
\usepackage{fancyhdr}       
\usepackage{graphicx}       
\graphicspath{{media/}}     
\usepackage{lipsum}
\usepackage{subcaption} 
\usepackage{authblk} 
\usepackage{amsmath}
\title{Design Choices That Matter: A Functional ANOVA Analysis for Remote Sensing Multi-Label Classification}

\author{
Maryam Gholami Shiri$^{1,2}$,
Eva Tuba$^{3,4,5}$,
Sa\v{s}o D\v{z}eroski$^{1}$,
Tome Eftimov$^{2,3}$,
Ana Nikolikj$^{2,3}$\\

$^1$Department of Knowledge Technologies, Jožef Stefan Institute, Ljubljana, Slovenia\\
$^2$Jožef Stefan International Postgraduate School, Ljubljana, Slovenia\\
$^3$Computer Systems Department, Jožef Stefan Institute, Ljubljana, Slovenia\\
$^4$Trinity University, San Antonio, TX, USA\\
$^5$Singidunum University, Belgrade, Serbia\\

\texttt{maryam.gholami.shiri@ijs.si, saso.dzeroski@ijs.si}\\
\texttt{eva.tuba@ijs.si, tome.eftimov@ijs.si, ana.nikolikj@ijs.si}
}

\usepackage{float}
\begin{document}
\maketitle

\begin{abstract}

Benchmarking deep learning (DL) models for multi-label classification (MLC) of remote sensing images (RSI) typically yields rankings that do not generalize beyond the evaluated datasets. In this work, we move beyond rankings by employing functional analysis of variance (fANOVA) to systematically quantify the contributions of individual design choices and their interactions to performance variability. We conduct two empirical analyses covering 48 and 20 DL models, respectively, spanning design choices such as network architecture, fine-tuning strategy, learning strategy, and initialization. By applying fANOVA across seven MLC RSI datasets, we construct dataset meta-representations that capture design-choice sensitivity profiles. Hierarchical clustering of these meta-representations reveals that datasets naturally group according to how they respond to design decisions, with patterns strongly linked to intrinsic dataset properties such as scale, spatial resolution, and label space complexity. Our findings show that for large-scale datasets, fine-tuning strategy and architecture are dominant factors, while in data-limited regimes, initialization becomes decisive. For intermediate regimes, the interaction between architecture and learning strategy governs performance.

\end{abstract}


\section{Introduction}
Remote sensing image (RSI) analysis engages with Earth surface images captured by satellites and aerial platforms, supporting the monitoring of phenomena in meteorology~\cite{Zhang2018WeatherRS}, agriculture~\cite{Chlingaryan2018AgricultureRS}, and urban planning~\cite{Huang2018UrbanRS}. Since a single image typically contains multiple geographic elements with different semantic labels, multi-label classification (MLC)~\cite{bogatinovski2022comprehensive} naturally arises as a core learning task. In recent years, deep learning (DL) has become the dominant paradigm for MLC in RSI, significantly advancing the state of the art~\cite{chen2021remote}. These advances are driven by powerful architectures, diverse training paradigms (e.g., end-to-end models or feature extractors for traditional ML), and transfer learning strategies such as linear evaluation~\cite{alain2016understanding} and full fine-tuning~\cite{yosinski2014transferable}. While the wide range of experimental choices enables flexibility, it also raises the question of which approach is best suited for a given dataset. This has motivated extensive benchmarking efforts across multiple datasets, with the goal of identifying the most effective experimental design choices~\cite{dimitrovski2023current}.

Although benchmarking plays a central role in the evaluation of DL models, interpreting benchmarking results remains limited to model rankings. However, it is widely known that the performance of DL models is strongly influenced by multiple factors, including network architecture, training strategy, and transfer learning. Moreover, MLC dataset characteristics (i.e., meta-features) such as size, spatial resolution, and semantic complexity directly affect the effectiveness of the aforementioned experimental design choices. Understanding the contribution and interaction of these factors is essential for moving beyond simple rankings that cannot be generalized to new datasets, toward actionable insights.

\textbf{Our contribution:} \textit{i)} We apply functional ANOVA (fANOVA) to two MLC benchmarking scenarios spanning 48 and  20 DL models across seven RSI datasets, quantifying the contribution of individual design choices and their interactions with performance variability; \textit{ii)} We construct dataset meta-representations from fANOVA importance scores and show via hierarchical clustering that datasets naturally group by design-choice sensitivity rather than overall performance level; \textit{iii)} We demonstrate that the dominant design factor shifts with dataset regime: fine-tuning drives large-scale settings, initialization governs data-limited regimes, and interactions between modules characterize intermediate regimes; and \textit{iv)} We reframe conclusions from two established benchmarking  studies~\cite{stoimchev2023deep,dimitrovski2023current} as dataset-conditional rather than universal, providing actionable, dataset-aware guidelines for understanding the model design choices. The data and code required to reproduce the study are available at: \url{https://zenodo.org/records/19630021}.

\section{Related work}
This section introduces the necessary background on MLC, fANOVA, and its previous applications, and outlines key meta-features of MLC datasets.

\noindent\textbf{Multi-label classification:} 
Multi-label classification (MLC) is a supervised learning setting where each instance can be associated with multiple labels from a predefined set~\cite{Tsoumakas2010MLC,Zhang2014MLC}. Unlike binary or multi-class classification, labels may co-occur. Formally, given an input space $\mathcal{X}$ and a label set $\mathcal{C}={c_1,\dots,c_L}$, the goal is to learn a function $f:\mathcal{X}\rightarrow 2^{\mathcal{C}}$ that maps each instance to a subset of labels. A key challenge in MLC is modeling label correlations, such as semantically related labels that frequently appear together (e.g., water and wetlands).

\noindent\textbf{Functional ANOVA:}
fANOVA~\cite{sobol1993sensitivity} is an interpretation technique that decomposes the variance of a model’s output into contributions from individual features and their interactions. It estimates feature importance through \textit{marginals}, i.e., the average model output when one feature is fixed while others vary; higher marginal variance indicates greater influence. Because exact computation is expensive, fANOVA is typically approximated using a random forest surrogate model that efficiently estimates these contributions~\cite{hutter2014efficient}.
fANOVA was originally introduced to analyze algorithm configuration spaces and identify influential hyperparameters in high-dimensional search spaces~\cite{hutter2014efficient}. Subsequent work extended the approach to multi-dataset settings, showing that hyperparameter importance can vary across datasets and provide useful priors for AutoML systems~\cite{vanrijn2018hyperparameter}. The method has since been integrated into hyperparameter optimization frameworks such as Optuna~\cite{akiba2019optuna} and applied to more complex scenarios, neural architecture analysis, and black-box optimization~\cite{klein2018reproducible,watanabe2023pedanova,nikolikj2025cmaes}. Motivated by these developments, we use fANOVA to analyze performance variability in MLC across datasets and experimental scenarios.

\noindent\textbf{Meta-features for MLC datasets:}
Prior work has explored the relationship between dataset characteristics and the performance of multi-label classification (MLC) methods through meta-learning analyses. In particular, Bogatinovski et al.~\cite{bogatinovski2022explaining} showed that meta-features describing the label space, such as label dependencies and distribution statistics, are especially informative for explaining performance differences across MLC methods. Their study also indicates that hyperparameter optimization yields only modest improvements relative to its computational cost. These findings highlight the importance of dataset properties for understanding MLC behavior and motivate further investigation into how dataset characteristics influence performance variability.

\section{Methodology}
We apply the extended fANOVA framework of Nikolikj et al.~\cite{nikolikj2025cmaes} to DL models benchmarking results across multiple MLC datasets to analyze the importance of design choices and their interactions. The analysis proceeds in two steps: (i) computing fANOVA importance scores to construct dataset meta-representations, and (ii) post-hoc analysis of the meta-representations to uncover patterns beyond traditional benchmarking rankings.

\noindent\textbf{Dataset meta-representations with fANOVA.}
Let $\mathcal{T}=\{D_k\}_{k=1}^{K}$ denote a set of MLC datasets and $\mathcal{M}=\{M_1,\dots,M_m\}$ a set of DL models. Each model $M_i$ is represented by an $n$-dimensional vector $\mathbf{x}_i$ describing key design choices (e.g., architecture, initialization, learning strategy, etc.). Evaluating $M_i$ on dataset $D_k$ produces a performance score $y_i$, forming $\mathcal{D}_k=\{(\mathbf{x}_i,y_i)\}_{i=1}^{m}$. For each $\mathcal{D}_k$, fANOVA decomposes model performance into contributions from individual design choices and their interactions. The resulting importance scores are aggregated into a dataset meta-representation $\boldsymbol{\phi}_k$, capturing how design choices influence performance on dataset $D_k$.

\noindent\textbf{Post-hoc analysis.}
We cluster the dataset meta-representations $\{\boldsymbol{\phi}_k\}_{k=1}^{K}$ to identify groups of datasets with similar design choice importance patterns, rather than performance. We then relate these patterns to dataset meta-features to understand how intrinsic dataset properties influence design choice importance.

\noindent\textbf{Why fANOVA?}
Several methods exist for attributing model performance to input features, such as permutation importance~\cite{breiman2001random}, SHAP  values~\cite{lundberg2017unified}, and marginal contribution analysis~\cite{owen2014sobol}. We choose fANOVA as it provides a globally consistent variance decomposition across the entire configuration space (unlike local methods such as SHAP), explicitly models higher-order interactions central to our analysis, and its random forest surrogate is tractable for pool sizes $m \leq 48$~\cite{hutter2014efficient}.


\section{Experimental Design}
We continue by explaining the experimental setup of our study, including the benchmarking data, the definition of modules representing model design choices, and the steps of the extended fANOVA analysis, along with the clustering procedure applied to the resulting meta-representations.

\noindent\textbf{Benchmarking scenarios:}  We consider two benchmarking scenarios reflecting practical model-selection settings previously studied for MLC in RSI.

\noindent\textit{Scenario 1: End-to-End DL vs. DL Feature Extraction.}
This scenario examines whether DL models are more effective when used as end-to-end predictors or as feature extractors combined with classical tree-based models. The experiment is based on the benchmarking study of Stoimchev et al.~\cite{stoimchev2023deep} and focuses on one family of DL models, convolutional neural networks (CNNs). It analyzes the impact of design choices such as network architecture and fine-tuning strategy and evaluates whether CNN-based feature extraction combined with tree ensemble methods can rival or outperform end-to-end CNN models. 

\noindent\textit{Scenario 2: DL Architecture and Initialization Strategies.}
This scenario investigates how the choice of DL model family and initialization strategy contributes to predictive performance. The experiment builds on the benchmarking study of Dimitrovski et al.~\cite{dimitrovski2023current} and compares a broader range of DL model families, including CNNs and transformer-based models, under different initialization strategies. 

\noindent\textbf{Design choices:} Depending on the benchmarking scenario, different sets of design choices are considered.
 In Scenario~1, the design choices from~\cite{stoimchev2023deep} are mapped to three modules representing DL models: \textit{network architecture}, \textit{fine-tuning strategy}, and \textit{learning strategy}, summarized in Table~\ref{tbl:exp1_modules}. Their combinations result in 48 DL models. In Scenario~2, the design choices from~\cite{dimitrovski2023current} are mapped to two modules: \textit{network architecture} and \textit{initialization strategy}, summarized in Table~\ref{tbl:exp2_modules}, yielding 20 deep MLC models.

\begin{table}[t]
\scriptsize
\centering
\caption{Design choices describing the DL models in Scenario~1}
\label{tbl:exp1_modules}
\begin{tabularx}{\textwidth}{p{3cm} X p{4cm}}
\hline
\textbf{Module} & \textbf{Description} & \textbf{Options} \\
\hline
Network architecture (Arch) & Defines the network structure: the layer number, size and type, and how they are interconnected. & VGG-16, VGG-19 \cite{Simonyan2015VGG}, ResNet-34, ResNet-50, ResNet-152 \cite{He2016ResNet}, EfficientNet-B0, EfficientNet-B1, EfficientNet-B2 \cite{Tan2019EfficientNet} \\
\hline
Fine-tuning strategy (FT) & Defines how ImageNet pretrained weights are updated: \textit{pre-trained} freezes all layers except the last fully connected layer; \textit{fine-tuned} updates all layers. & pre-trained, fine-tuned \\ 
\hline
Learning strategy (LS) & Defines how the model is used for inference: \textit{end-to-end} produces predictions directly via the last fully connected layer; \textit{feature extraction} uses the model as a feature extractor for a separate downstream model. & end-to-end, Random Forest~\cite{Breiman2001RF}, Extra Trees~\cite{Geurts2006ExtraTrees}\\
\hline
\end{tabularx}
\end{table}

\begin{table}[t]
\scriptsize
\centering
\caption{Design choices describing the DL models in Scenario~2.}
\label{tbl:exp2_modules}
\begin{tabularx}{\textwidth}{p{3cm} X p{4cm}}
\hline
\textbf{Module} & \textbf{Description} & \textbf{Options} \\
\hline
Network architecture (Arch) & Defines the network structure: the layer number, size and type, and how they are interconnected & AlexNet \cite{Krizhevsky2012AlexNet}, VGG-16 \cite{Simonyan2014VGG}, ResNet-50, ResNet-152 \cite{He2016ResNet}, DenseNet-161 \cite{Huang2017DenseNet}, EfficientNet-B0 \cite{Tan2019EfficientNet}, ViT \cite{Dosovitskiy2020ViT}, MLPMixer \cite{Tolstikhin2021MLPMixer}, ConvNeXt \cite{Liu2022ConvNeXt}, SwinT \cite{Liu2021Swin}\\
\hline
Initialization strategy (Init) & Defines how the weights are initialized. Either using random initialization in case of $from scratch$ option, or pre-trained weights and then fine-tuned in case of $pretrained$. & from scratch, pretrained \\
\hline
\end{tabularx}
\end{table}

\noindent\textbf{MLC Benchmark Datasets for RSI:}
The first scenario considers seven publicly available RSI datasets that vary in key meta-features, including the number of images, label space dimensionality, image resolution, and label cardinality (average number of labels per image). The second scenario also includes seven RSI datasets. Table~\ref{tbl:datasets_combined} summarizes the datasets and their seven meta-features. Six datasets are shared between the two scenarios, while each includes one distinct dataset. The reported meta-features are a subset of those introduced in Bogatinovski et al.~\cite{bogatinovski2022explaining}. 
\textit{BigEarthNet-19} and \textit{BigEarthNet-43} represent very large-scale datasets with more than 590k samples, moderate label spaces (19 and 43 labels), relatively low label density, and a consistent spatial resolution of 10\,m with $256^2$ image size. \textit{MLRSNet} also belongs to the large-scale regime, containing over 100k samples, but differs by having the largest label space (60 labels) and a wider range of spatial resolutions (0.1--10\,m), reflecting higher variability in the imagery. \textit{PlanetUAS} represents a mid-to-large scale dataset with more than 40k samples and moderate spatial resolution (3\,m), but with lower label cardinality. In contrast, \textit{UCM}, \textit{AID}, and \textit{DFC-15} correspond to medium-scale datasets with a few thousand samples, moderate label spaces (8–17 labels), and high spatial resolution imagery (0.05–0.5\,m), although their image sizes differ. Finally, \textit{Ankara} represents a data-limited regime with only 216 samples, a large label space (29 labels), high label density, and small image size ($64^2$).

\begin{table}[t]
\caption{MLC datasets and meta-features. Cardinality - the average labels per image, density - the average proportion of images per label, and spatial resolution - the ground area represented by a pixel. $^\dagger$ Scenario~1 only; $^\ddagger$ Scenario~2 only.}
\label{tbl:datasets_combined}
\centering
\begin{adjustbox}{max width=\textwidth}
\scriptsize
\begin{tabular}{@{}llcccccc@{}}
\toprule
\textbf{Dataset} & \textbf{Type} & \textbf{\#Labels} & \textbf{Cardinality}
  & \textbf{Density$^\dagger$} & \textbf{\#Samples}
  & \textbf{Spatial Resolution$^\ddagger$} & \textbf{Image Size} \\
\midrule
Ankara$^\dagger$     & Aerial RGB & 29 & 9.12 & 0.536 & 216     & --        & $64^2$    \\
UCM                  & Aerial RGB & 17 & 3.33 & 0.476 & 2,100   & 0.3\,m    & $256^2$   \\
AID                  & Aerial RGB & 17 & 5.15 & 0.468 & 3,000   & 0.5--8\,m & $600^2$   \\
DFC-15               & Aerial RGB &  8 & 2.80 & 0.465 & 3,341   & 0.05\,m   & $600^2$   \\
PlanetUAS$^\ddagger$ & Aerial RGB & 17 & 2.90 & --    & 40,479  & 3\,m      & $256^2$   \\
MLRSNet              & Aerial RGB & 60 & 5.77 & 0.144 & 109,151 & 0.1--10\,m& $256^2$   \\
BigEarthNet-19       & Aerial RGB & 19 & 2.90 & 0.263 & 590,326 & 10\,m     & $256^2$   \\
BigEarthNet-43       & Aerial RGB & 43 & 2.97 & 0.247 & 590,326 & 10\,m     & $256^2$   \\
\bottomrule
\end{tabular}
\end{adjustbox}
\end{table}

\noindent\textbf{Performance data:} For Scenario~1, we reuse performance data 
from~\cite{stoimchev2023deep}, with the models evaluated via \textit{ranking loss} (the fraction of incorrectly ordered label pairs, ranging in $[0,1]$, where lower is better). For Scenario~2, we reuse performance data from~\cite{dimitrovski2023current}, with models evaluated via \textit{mean average precision} (mAP) (how well relevant labels are ranked ahead of irrelevant ones, ranging in $[0,100]$, where higher is better). Since the scenarios use different metrics, fANOVA importance score comparisons are made \emph{within} each scenario only.

\noindent\textbf{Dataset meta-representations}: Applying fANOVA to each of the seven datasets yields meta-representations that capture the importance profiles of the design choices. For Scenario~1, where models are described by three modules, the dimensionality of the meta-representation is 
$3 + \binom{3}{2} + \binom{3}{3} = 7$, 
corresponding to the main effects, pairwise interactions, and the three-way interaction. For Scenario~2, where models are defined by two modules, the meta-representations have dimensionality 
$2 + \binom{2}{2} = 3$, 
representing the main effects and their interaction.

\noindent\textbf{Clustering of dataset meta-representations:} Dataset meta-representations are clustered using Hierarchical Clustering (HC)~\cite{Mllner2011ModernHA} and clusters selected using the dendrogram. 
Clustering quality is evaluated using Silhouette coefficient (cluster cohesion and separation, range [-1,1]) and cophenetic correlation (dendrogram fidelity to pairwise distances, range [-1,1]).

\begin{figure*}[!ht]
    \centering
    \begin{subfigure}{0.49\textwidth}
        \centering
        \includegraphics[width=.82\textwidth]{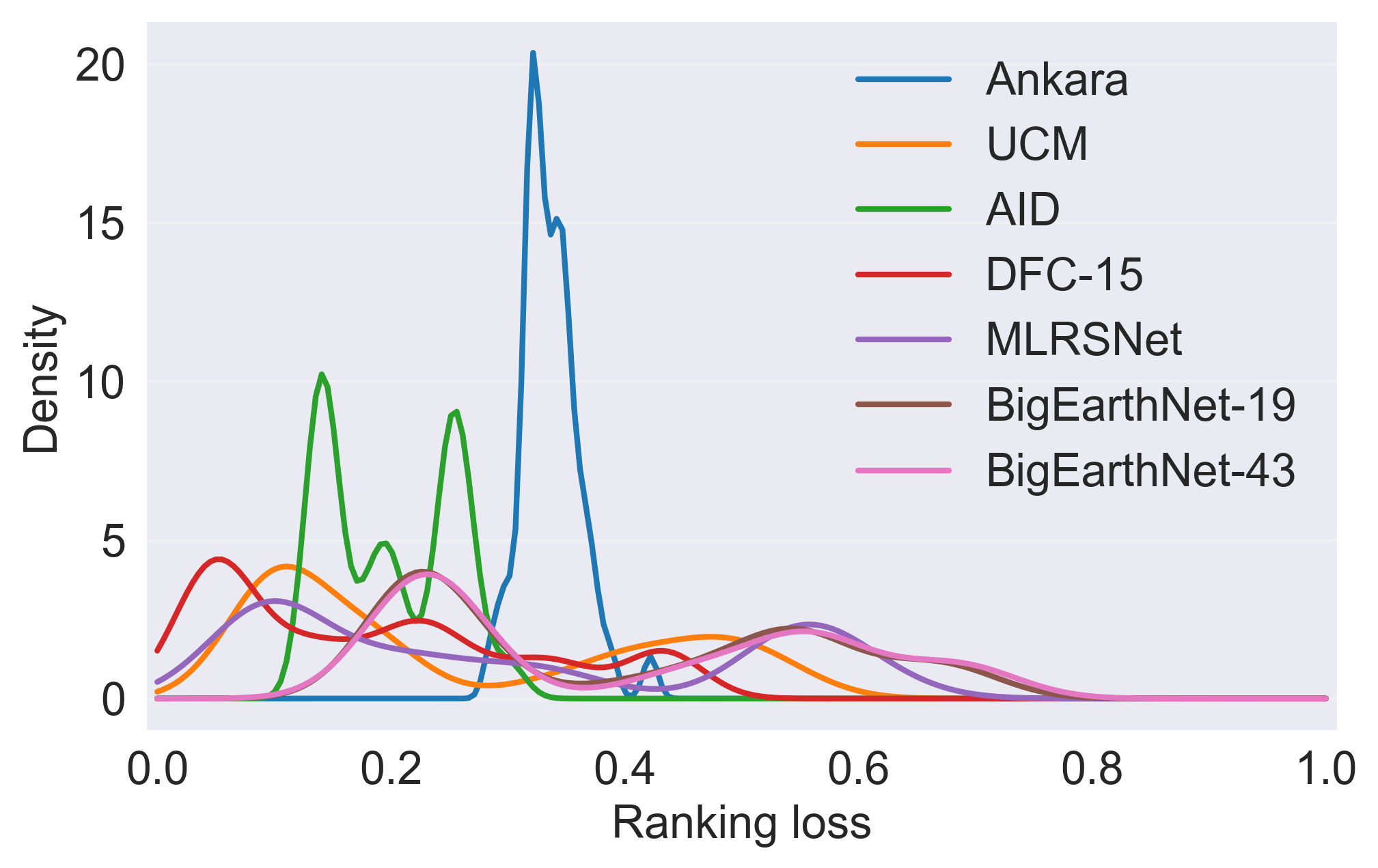}
        \caption{}
        \label{fig:exp1_kde}
    \end{subfigure}
    \begin{subfigure}{0.49\textwidth}
        \centering
        \includegraphics[width=.82\textwidth]{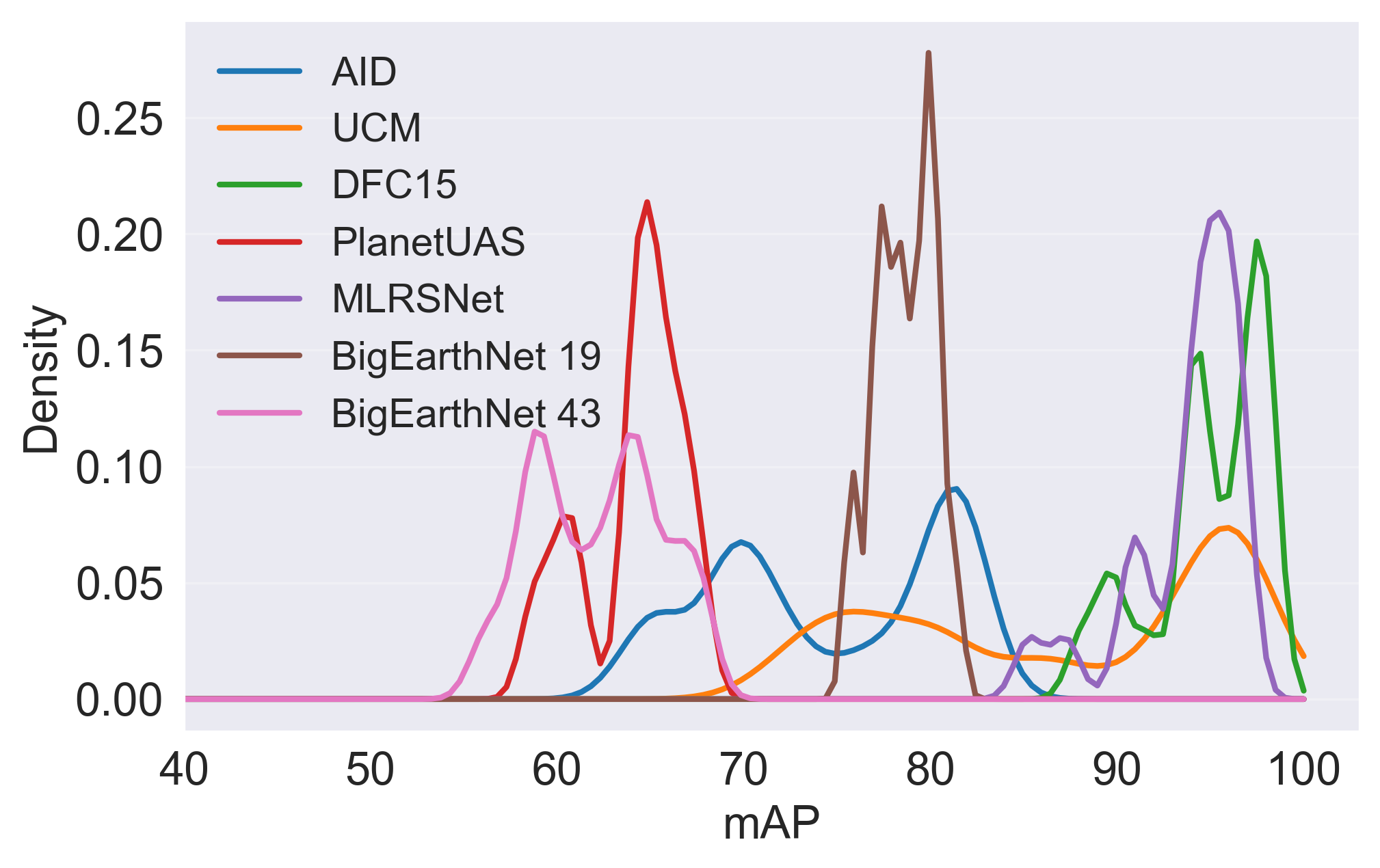}
        \caption{}
        \label{fig:exp2_kde}
    \end{subfigure}
   \caption{Kernel density estimates (KDEs) of (a) ranking loss in Scenario~1 (lower is better) and (b) mAP in Scenario~2 (higher is better) across datasets.}
\label{fig:raw_perf_var}
\end{figure*}

\section{Results}
We first visualize the raw performance variability. Next, cluster fANOVA-based dataset meta-representations to identify datasets with similar importance patterns and finally relate these patterns to dataset meta-features. 


\noindent\textbf{Variability of raw performance data.}
\textbf{Scenario~1:} In Figure~\ref{fig:exp1_kde}, \textit{Ankara} and \textit{AID} show tight ranking-loss distributions concentrated at low values, indicating limited sensitivity to design choices. For \textit{UCM}, \textit{DFC-15}, \textit{MLRSNet}, and both \textit{BigEarthNet} the distributions are wide, spanning a large portion of the $[0,1]$ interval, suggesting stronger sensitivity. \textbf{Scenario~2:} Most datasets show low mAP variability through the tight distributions (Figure~\ref{fig:exp2_kde}). For \textit{DFC-15} and \textit{MLRSNet}, the distributions concentrate at high mAP values, whereas for \textit{PlanetUAS} and both \textit{BigEarthNet} at moderate values, suggesting low sensitivity to design choices. In contrast, \textit{AID} and \textit{UCM} exhibit higher variability, indicating greater sensitivity to design choices.

\begin{figure}[!ht]
    \centering
    \begin{subfigure}{0.49\textwidth}
        \centering
        \includegraphics[width=.82\textwidth]{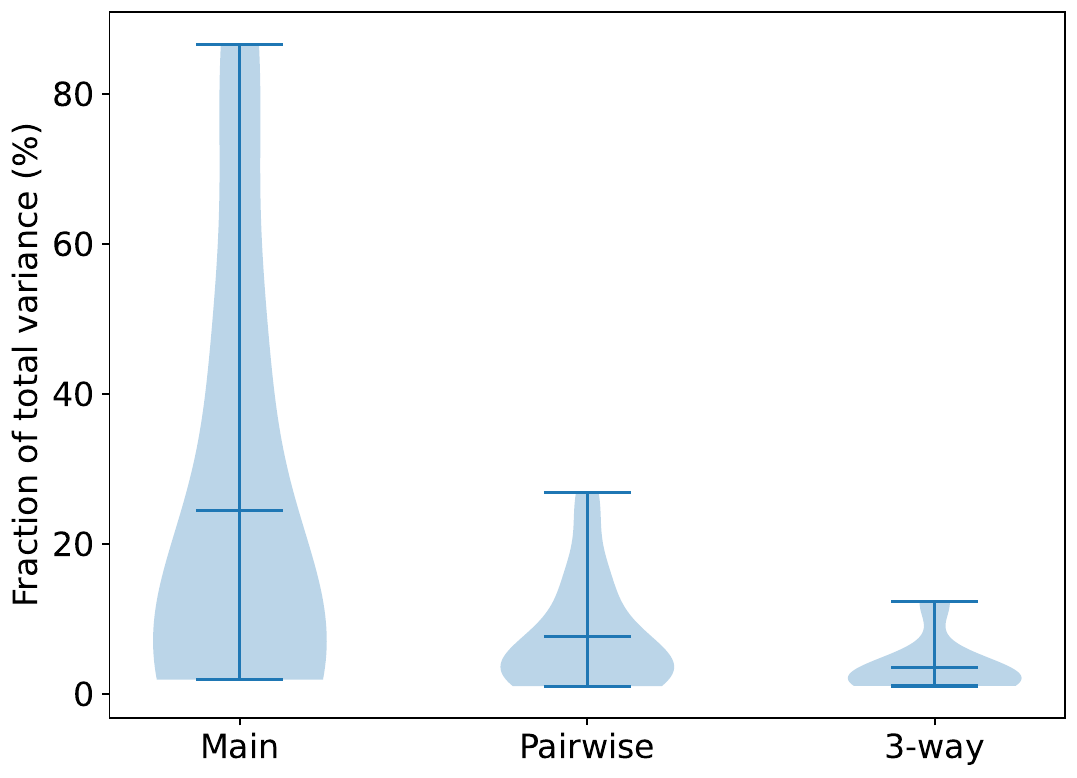}
        \caption{}
        \label{fig:imp-effect_types_exp1}
    \end{subfigure}
    \begin{subfigure}{0.49\textwidth}
        \centering
        \includegraphics[width=.82\textwidth]{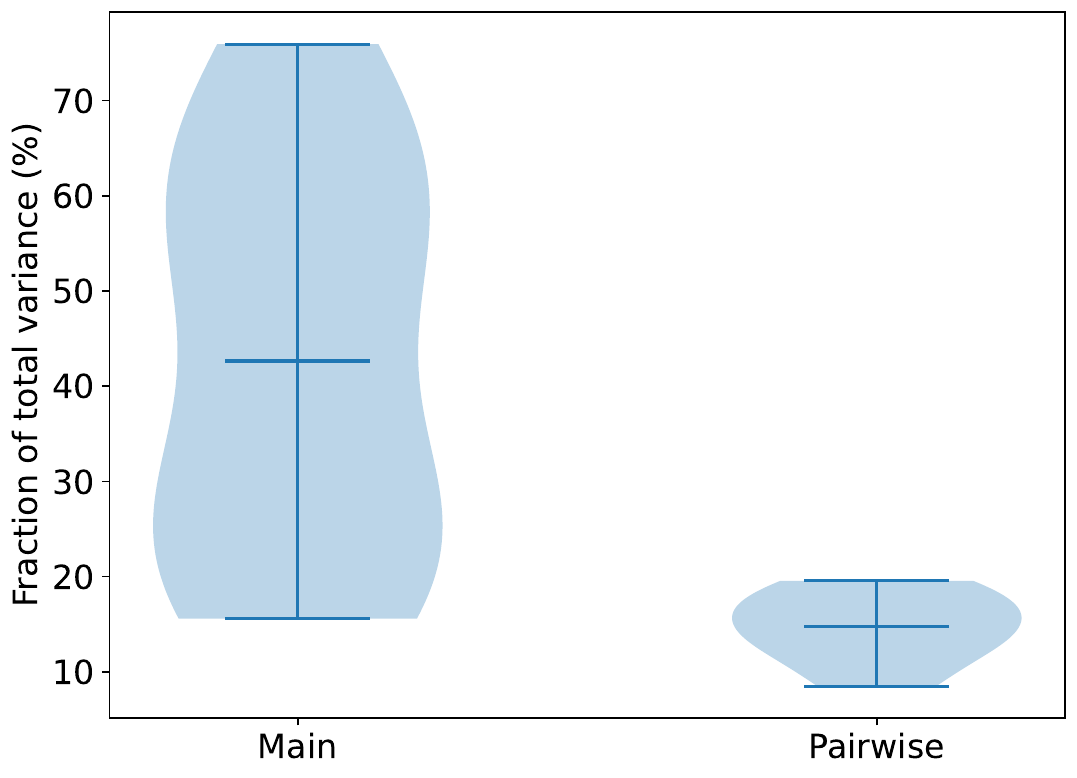}
        \caption{}
        \label{fig:imp-effect_types_exp2}
    \end{subfigure}
    \caption{Violin plots of variance distributions for individual and interaction effects of modules across the seven datasets in (a) Scenario~1 and (b) Scenario~2.}
    \label{fig:imp-effect_types}
\end{figure}

\noindent\textbf{Contribution of Design Choices and Their Interactions:} We analyze the contribution of individual design choices and their interactions to performance variability. This reveals whether design choices influence performance independently or mainly through their combinations. The quality of the fANOVA surrogate model is assessed using the in-sample coefficient of determination $R^2$, with median of 0.966 in Scenario 1 and 0.831 in Scenario 2, indicating a good fit and supporting the reliability of the resulting fANOVA importance estimates. \textbf{Scenario~1:} Individual effects dominate, with a heavy tail toward the high variance portion (Figure~\ref{fig:imp-effect_types_exp1}) and a single design choice explains more than 80\% of the variance in some cases. Pairwise interactions contribute substantially less, and triplet interactions are near zero, indicating that modules largely contribute independently to performance. \textbf{Scenario~2:} Individual effects again dominate, with several cases exceeding 70\% of total variance and a median around 40\% (Figure~\ref{fig:imp-effect_types_exp2}). Pairwise interactions remain non-negligible (8--20\%), indicating that the combination of design choices can still substantially influence final performance.

\noindent\textbf{Clustering of Dataset Meta-representations (Design-Choice Sensitivity):} Here, we analyze similarities between MLC datasets with respect to the importance of design-choice. Hierarchical clustering using cosine distance and average linkage yielded silhouette scores of 0.7571 and 0.5343, with cophenetic correlation of 0.9674 and 0.8016 for Scenarios 1 and 2, respectively.

\begin{figure}[t]
    \centering
    \begin{subfigure}{0.49\textwidth}
        \centering
        \includegraphics[width=.82\textwidth]{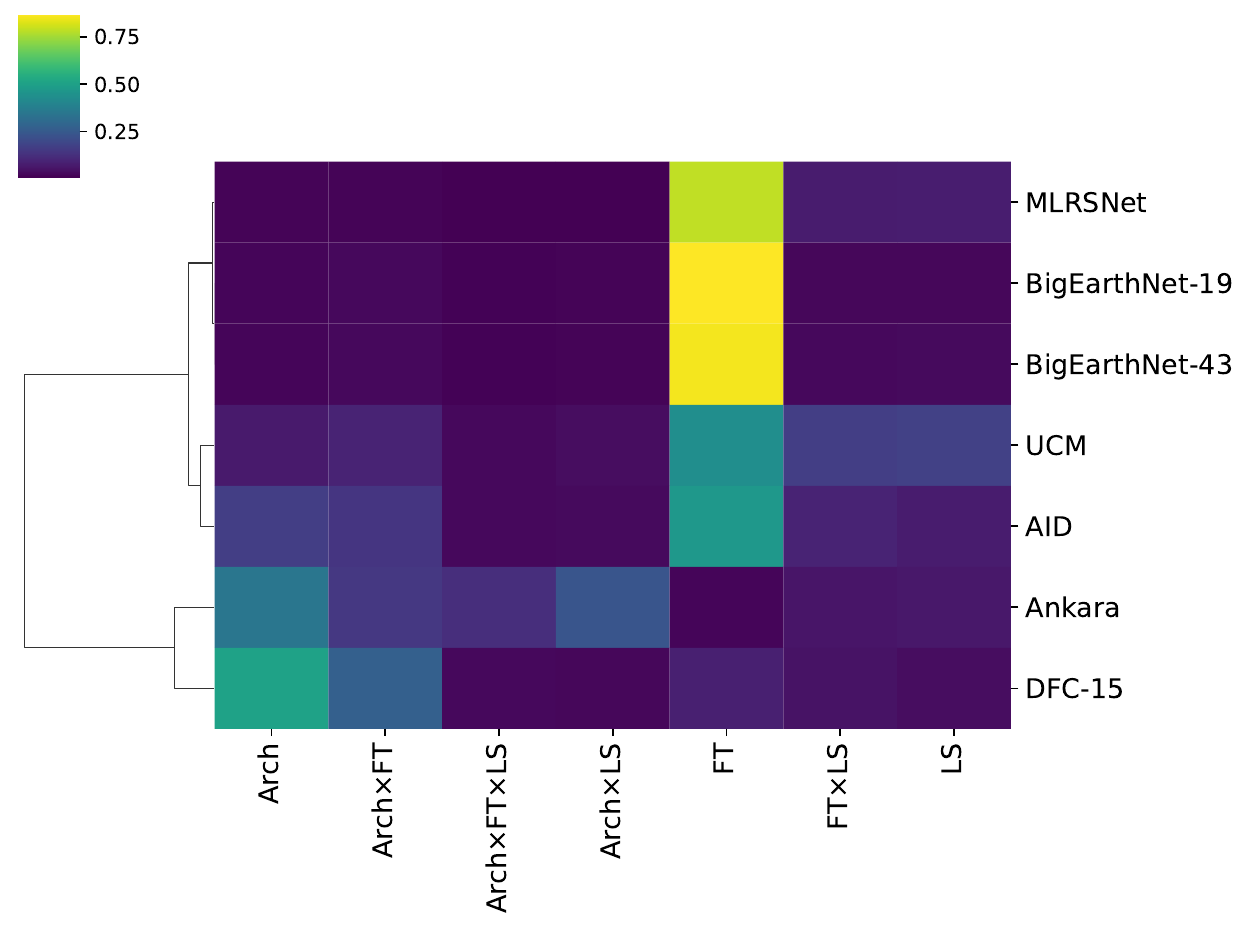}
        \caption{}
        \label{fig:exp1_clustermap}
    \end{subfigure}
    \begin{subfigure}{0.49\textwidth}
        \centering
        \includegraphics[width=.82\textwidth]{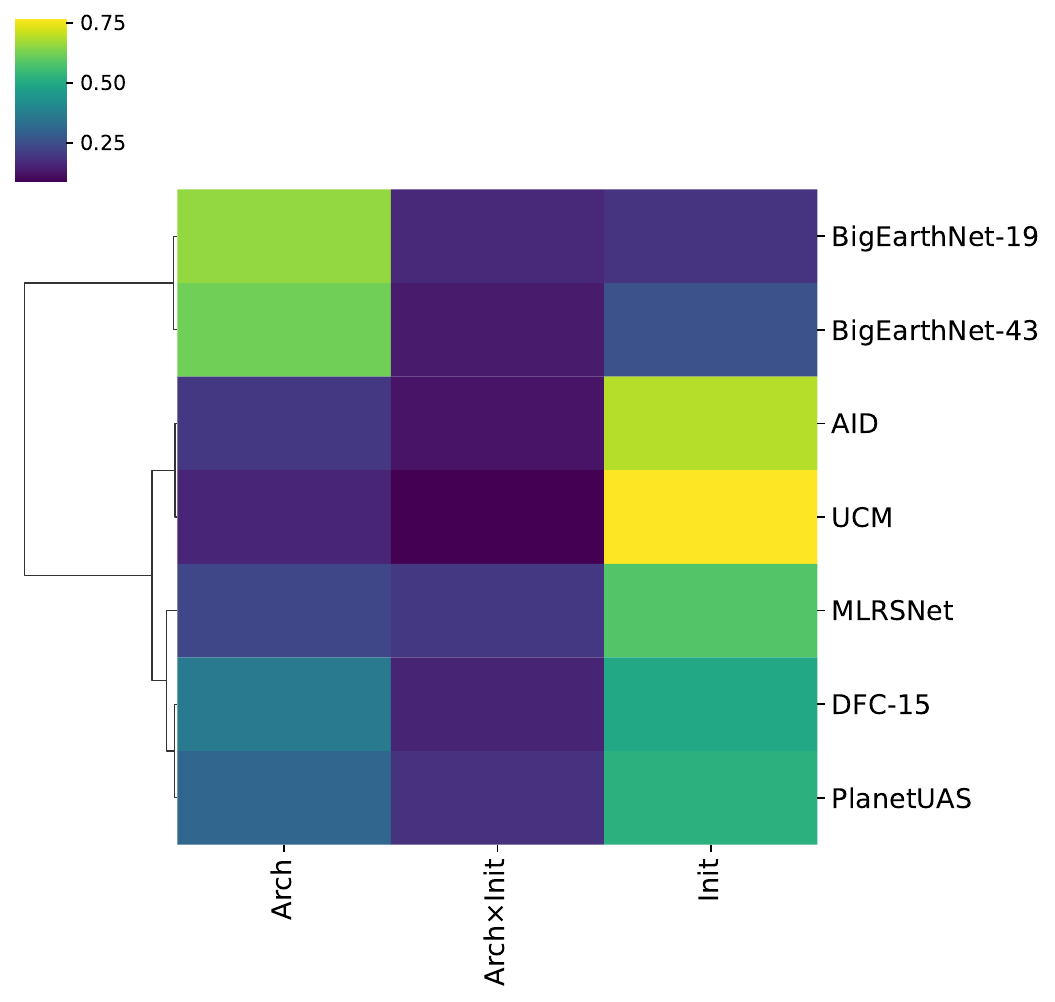}
        \caption{}
        \label{fig:exp2_clustermap}
    \end{subfigure}
    \caption{Clustermap showing hierarchical clustering of MLC dataset based on module effects, with rows as MLC datasets, columns as module effects, and color intensity as variance portion (yellow = higher) (a) Scenario~1 and (b) Scenario~2.}
    \label{fig:clustermap}
\end{figure}

\begin{figure*}[t]
    \centering
    \begin{subfigure}[t]{0.32\textwidth}
        \centering
        \includegraphics[width=\linewidth]{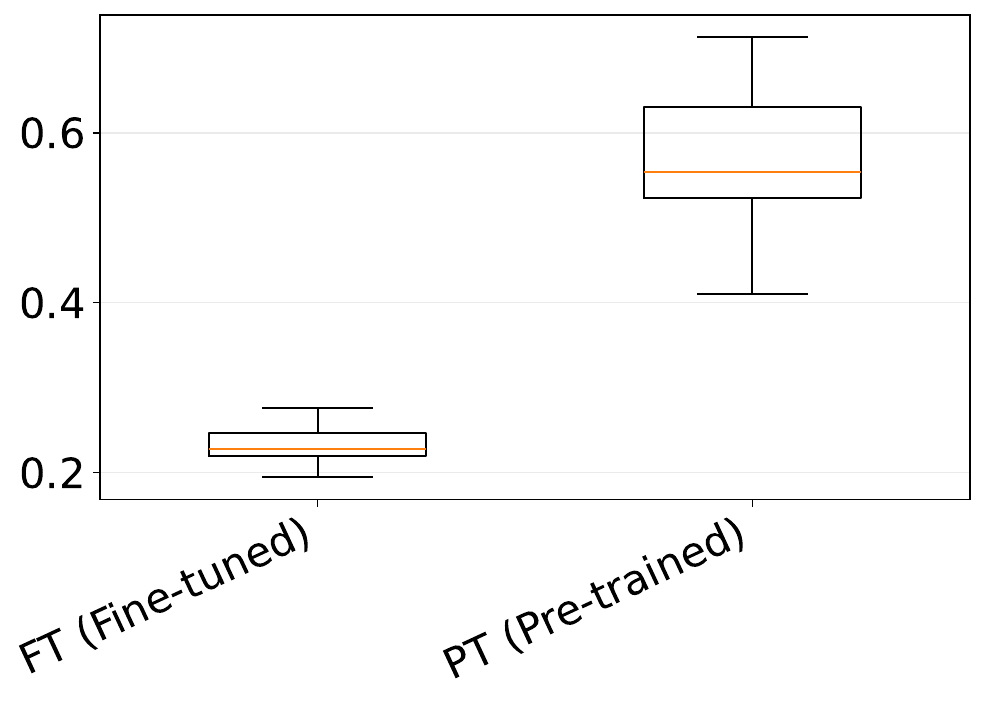}
        \caption{Fine-tuning}
        \label{fig:marginal_ft_ben43}
    \end{subfigure}
    \begin{subfigure}[t]{0.32\textwidth}
        \centering
        \includegraphics[width=\linewidth]{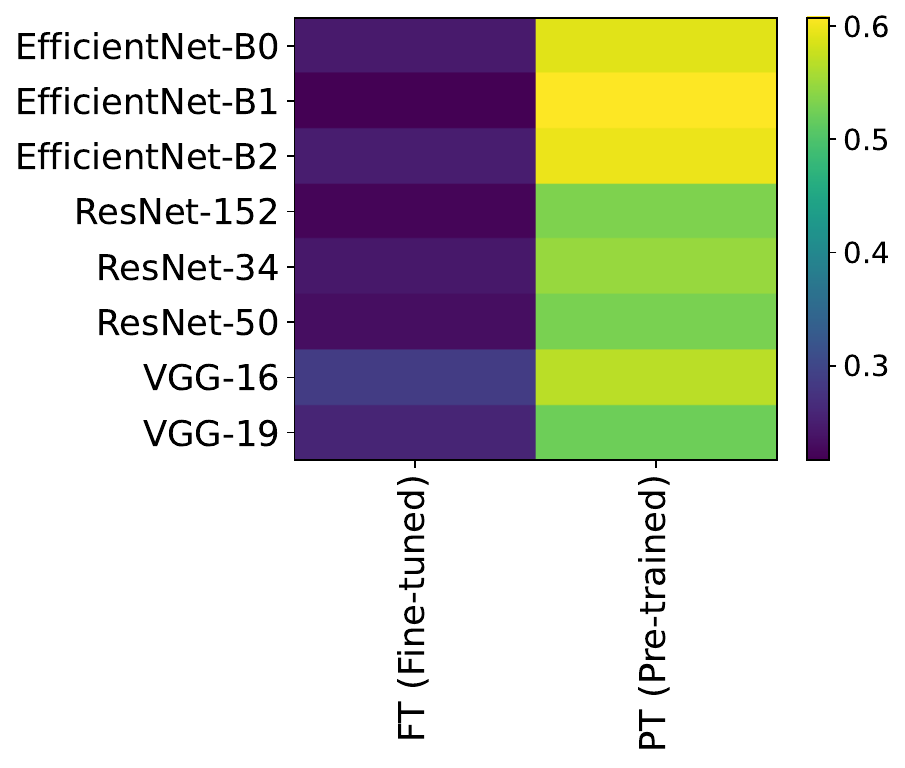}
        \caption{Architecture $\times$ Fine-tuning}
        \label{fig:marginal_ft-arch_ben43}
    \end{subfigure}
    \caption{Marginal effects of modules for \textit{BigEarthNet-43} (Cluster 1) in Scenario 1.}
    \label{fig:marginals_ben43}
\end{figure*}

\begin{figure*}[!t]
    \centering
    \begin{subfigure}[t]{0.32\textwidth}
        \centering
        \includegraphics[width=\linewidth]{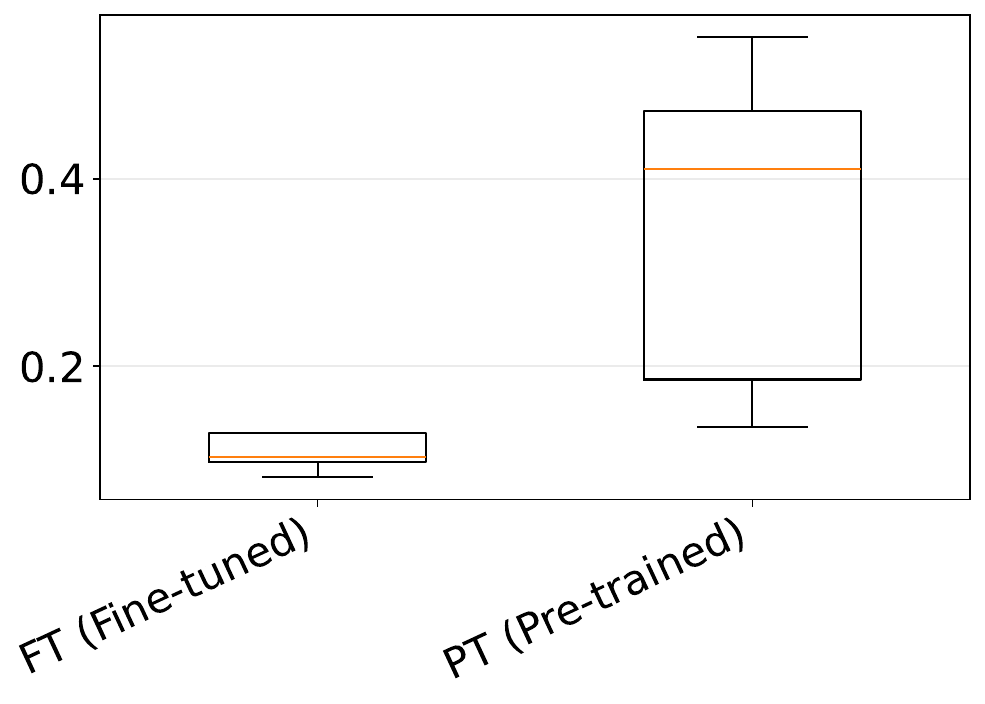}
        \caption{Fine-tuning strategy}
        \label{fig:marginal_ft_ucm}
    \end{subfigure}
    \begin{subfigure}[t]{0.32\textwidth}
        \centering
        \includegraphics[width=\linewidth]{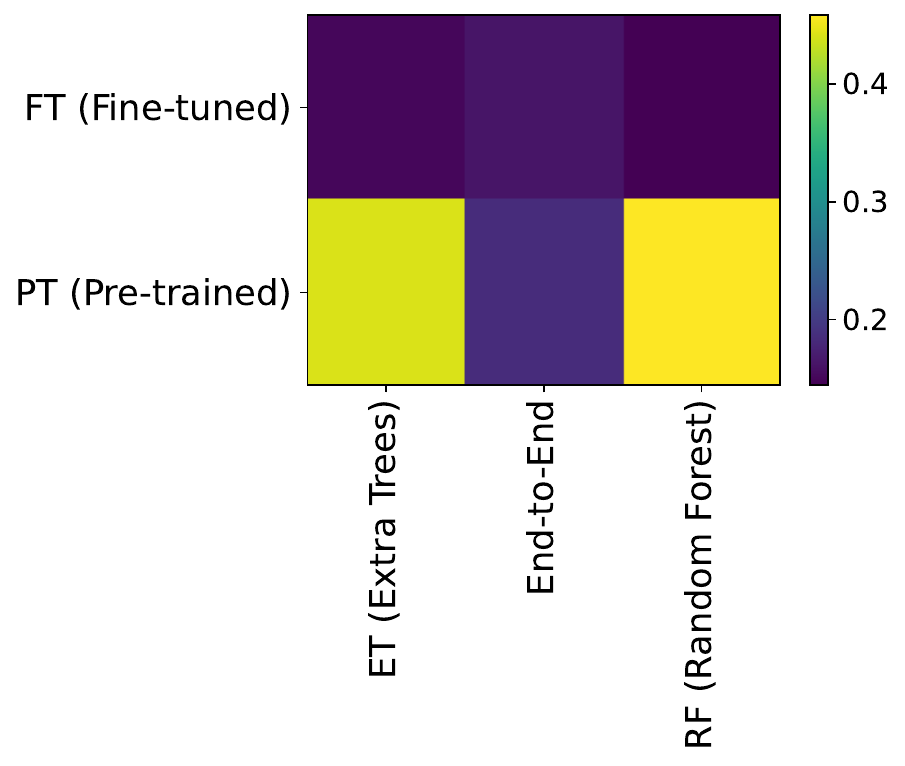}
        \caption{Fine-tuning $\times$ Learning strategy}
        \label{fig:marginal_ft_ls_ucm}
    \end{subfigure}
    \caption{Marginal effects of modules for \textit{UCM} (Cluster 2) in Scenario 1.}
    \label{fig:marginals_ucm}
\end{figure*}

\begin{figure*}[!t]
    \centering
    \begin{subfigure}[t]{0.32\textwidth}
        \centering
        \includegraphics[width=\linewidth]{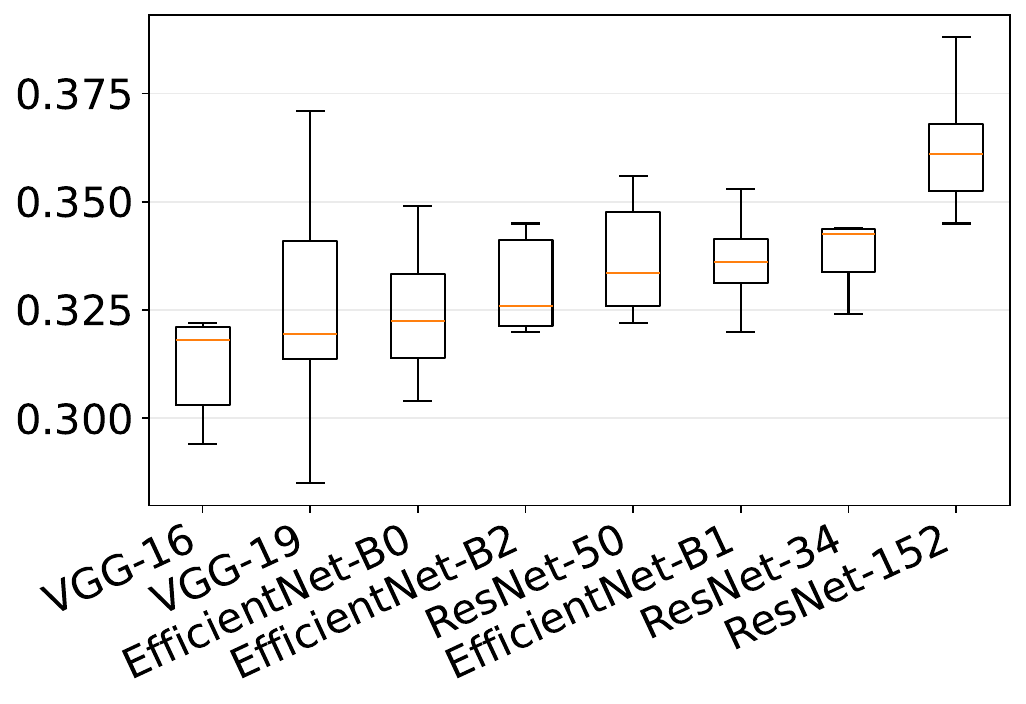}
        \caption{Architecture}
        \label{fig:marginal_arch_ankara}
    \end{subfigure}
    \begin{subfigure}[t]{0.32\textwidth}
        \centering
        \includegraphics[width=\linewidth]{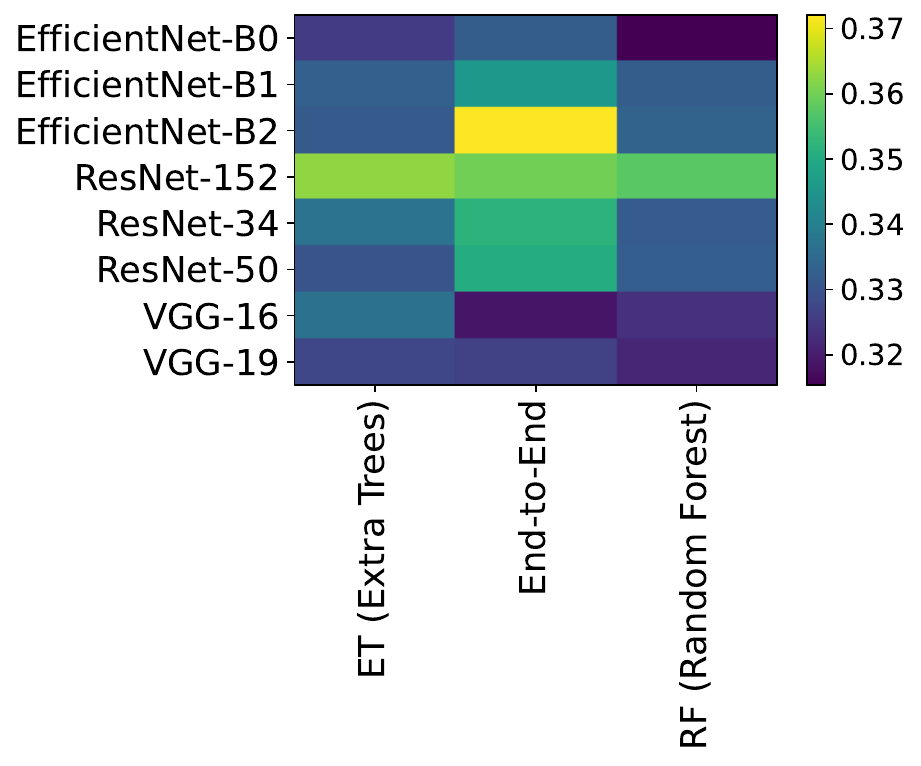}
        \caption{Architecture $\times$ Learning strategy}
        \label{fig:marginal_arch_ls_ankara}
    \end{subfigure}

    \caption{Marginal effects of modules for \textit{Ankara} (Cluster 3) in Scenario 1.}
    \label{fig:marginals_ankara}
\end{figure*}

\begin{figure*}[!ht]
    \centering
    \begin{subfigure}[t]{0.32\textwidth}
        \centering
        \includegraphics[width=\linewidth]{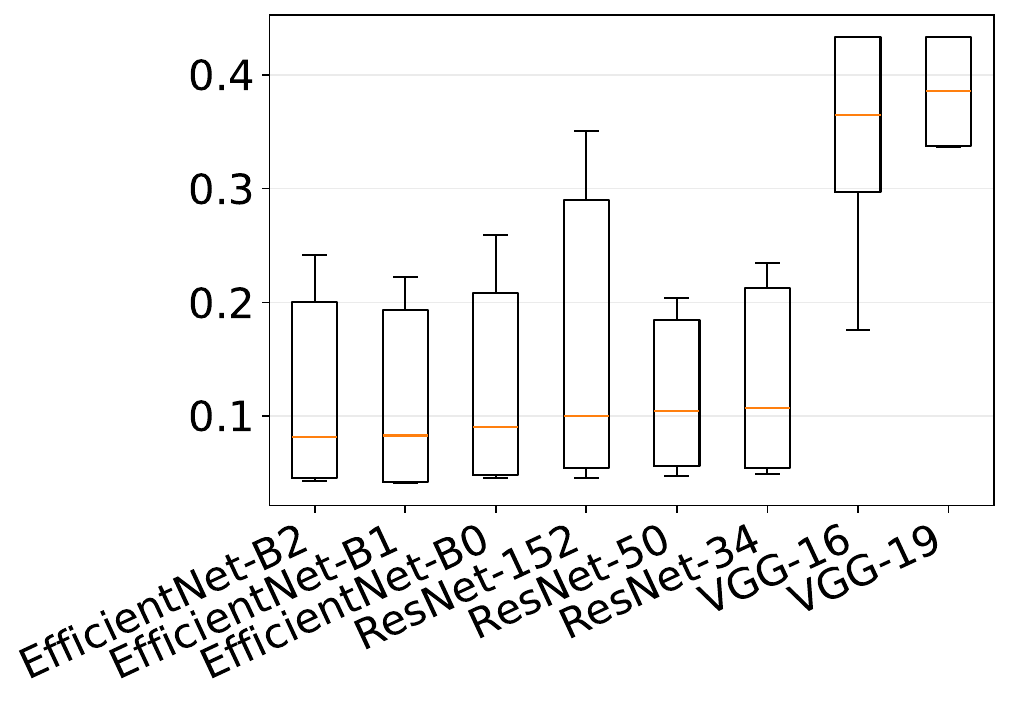}
        \caption{Architecture}
        \label{fig:marginal_arch_dfc}
    \end{subfigure}
    \begin{subfigure}[t]{0.32\textwidth}
        \centering
        \includegraphics[width=\linewidth]{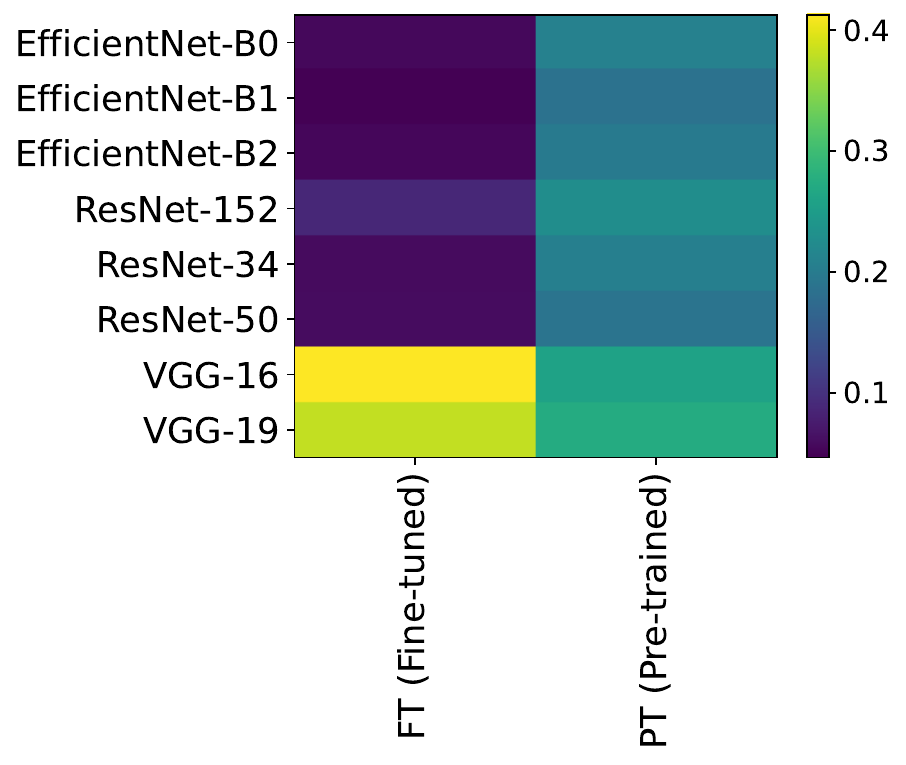}
        \caption{Architecture $\times$ Fine-tuning Strategy}
        \label{fig:marginal_archxft_dfc}
    \end{subfigure}
    \caption{Marginal effects of modules for the \textit{DFC-15} (Cluster 3) in Scenario 1.}
    \label{fig:marginals_dfc}
\end{figure*}

\noindent\textbf{Scenario 1:} Figure~\ref{fig:exp1_clustermap} presents the clustering of dataset meta-representations, where rows correspond to datasets and columns represent the variance contributions of individual modules and their interactions (e.g., Module$_1 \times$ Module$_2$). Color intensity indicates the contribution magnitude (yellow = higher). Three distinct clusters emerge, with clear differences in design-choice sensitivity despite the small number of datasets. \noindent\textbf{Cluster~1 (\textit{BigEarthNet-19}, \textit{BigEarthNet-43}, \textit{MLRSNet}).} The fine-tuning strategy dominates, indicating that adapting pre-trained representations to the target data is crucial for achieving good performance on large-scale datasets. Fine-tuned models consistently achieve lower ranking loss than those using pre-trained weights (Figure~\ref{fig:marginal_ft_ben43}), and the architecture$\times$fine-tuning interaction is negligible, indicating that fine-tuning improves performance uniformly across architectures (Figure~\ref{fig:marginal_ft-arch_ben43}). Similar observations hold for the learning strategy. \noindent\textbf{Cluster~2 (\textit{UCM}, \textit{AID}).} For the medium-scale datasets, the importance of fine-tuning decreases while other design choices become more influential, suggesting that performance depends more on their combination than on a single dominant factor. Fine-tuned models still outperform pre-trained ones (Figure~\ref{fig:marginal_ft_ucm}). The learning strategy has little impact once representations are adapted to the dataset with fine-tuning (Figure~\ref{fig:marginal_ft_ls_ucm}). In contrast, without fine-tuning, end-to-end approaches outperform classical classifiers such as Random Forest and Extra Trees.
\noindent\textbf{Cluster~3 (\textit{Ankara}, \textit{DFC-15}).} The architecture module becomes more influential, and its interaction with fine-tuning increases, indicating that performance depends more strongly on the joint selection of the representation backbone and learning strategy. This effect is evident for \textit{Ankara}, a small, low-resolution dataset, where simpler architectures such as VGG outperform deeper models like ResNet, suggesting that model complexity should match dataset scale (Figure~\ref{fig:marginal_arch_ankara}).The interaction with the learning strategy reveals that VGG-16 combined with an end-to-end approach achieves the best performance (Figure~\ref{fig:marginal_arch_ls_ankara}), indicating that joint representation learning can compensate for simpler architectures. For \textit{DFC-15}, the architecture becomes critical due to the high spatial resolution of the images. Modern architectures such as EfficientNet and ResNet outperform older architectures like VGG (Figure~\ref{fig:marginal_arch_dfc}). These models are better in capturing multi-scale and high-resolution patterns through deeper structures and improved receptive fields. The interaction between architecture and fine-tuning is pronounced (Figure~\ref{fig:marginal_archxft_dfc}), with fine-tuning benefiting some architectures more than others and even being detrimental for VGG models.

\begin{figure*}[t]
    \centering
    \begin{subfigure}[t]{0.32\textwidth}
        \centering
        \includegraphics[width=\linewidth]{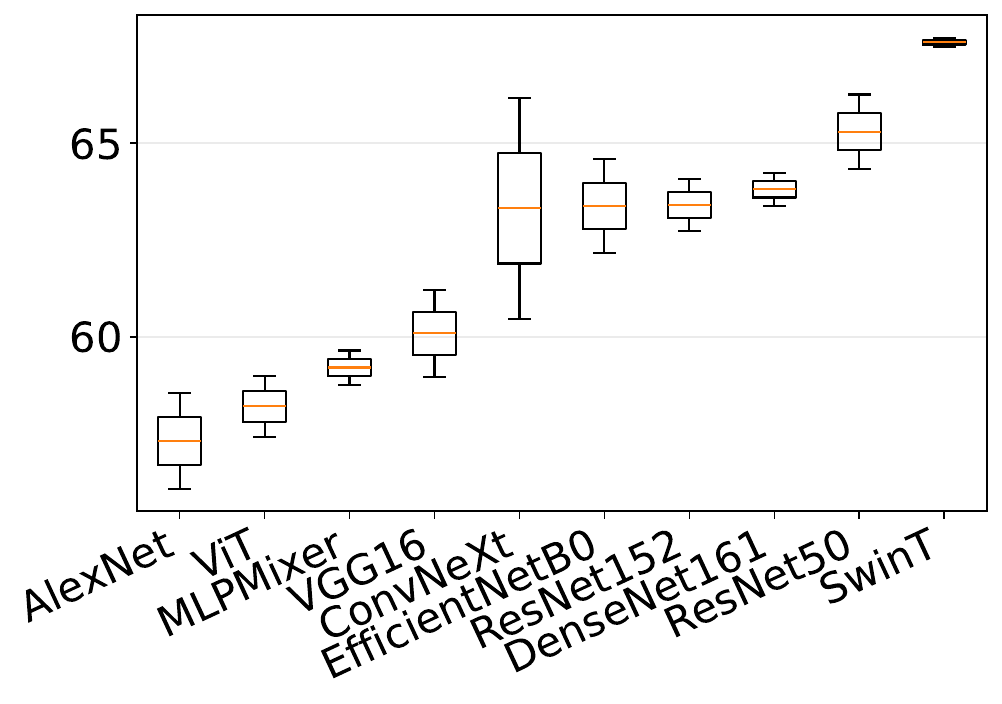}
        \caption{Architecture}
        \label{fig:exp2_marginal_arch_ben43}
    \end{subfigure}
    \hfill
    \begin{subfigure}[t]{0.32\textwidth}
        \centering
        \includegraphics[width=\linewidth]{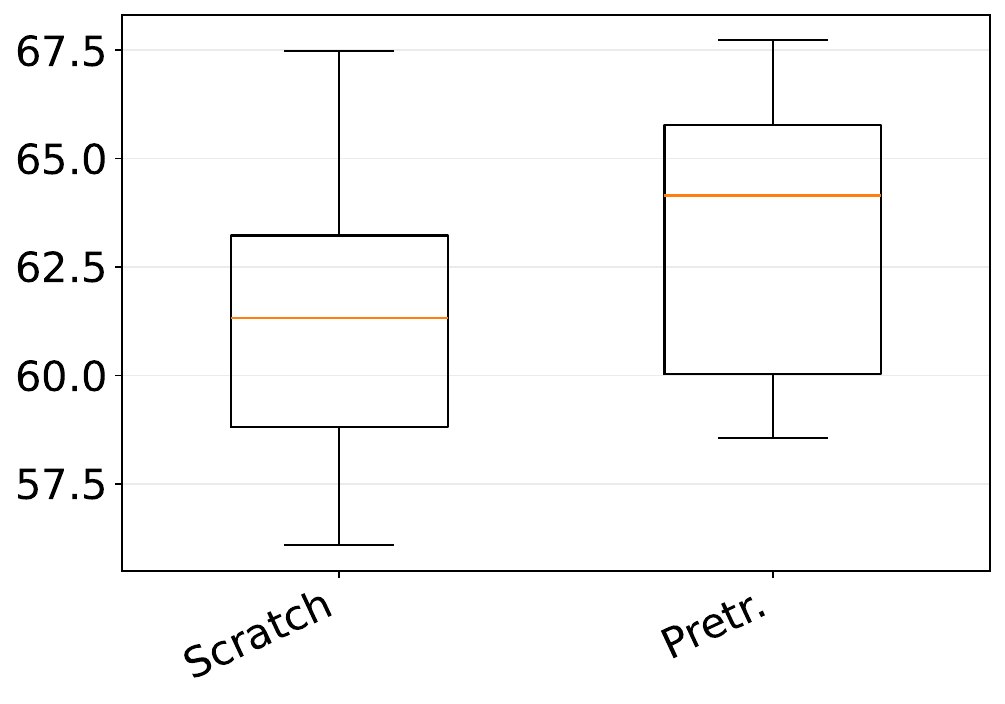}
        \caption{Initialization strategy}
        \label{fig:exp2_marginal_init_ben43}
    \end{subfigure}
    \hfill
    \begin{subfigure}[t]{0.32\textwidth}
        \centering
        \includegraphics[width=\linewidth]{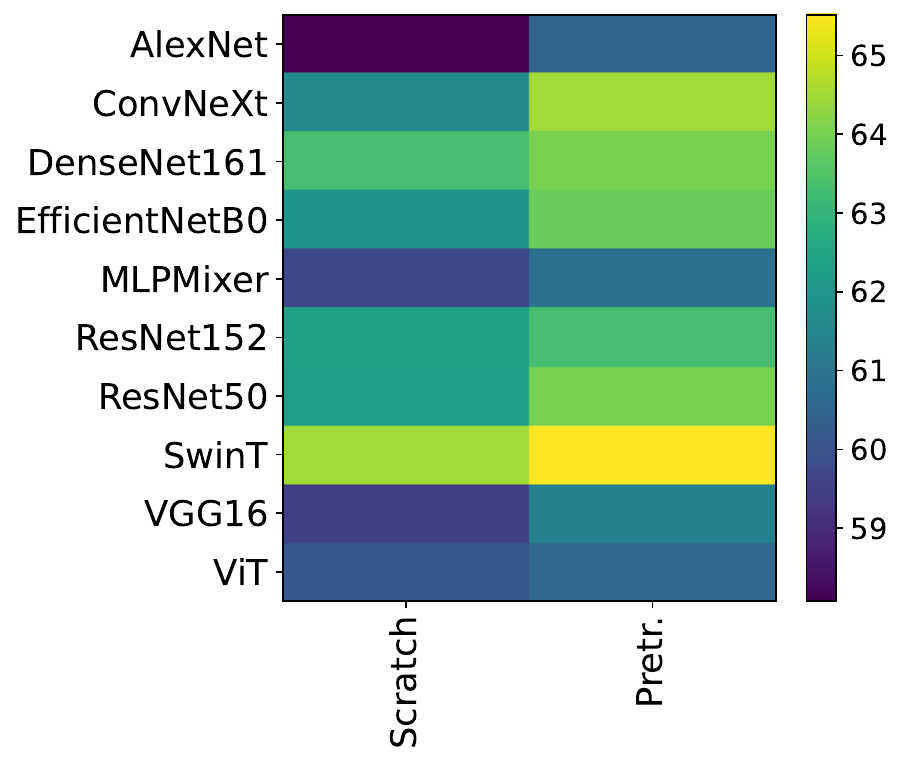}
        \caption{Architecture $\times$ Initialization}
        \label{fig:exp2_marginal_arch_init_ben43}
    \end{subfigure}

    \caption{Marginal effects of modules for\textit{BigEarthNet-43} (Cluster 1) in Scenario 2.}
    \label{fig:exp2_marginals_ben43}
\end{figure*}

\noindent\textbf{Scenario 2:} The results confirm that no single module is universally dominant across all datasets, even when only two modules are considered. Three groups emerge, revealing latent similarities in design-choice sensitivity across datasets (Figure~\ref{fig:exp2_clustermap}). \noindent\textbf{Cluster 1 (\textit{BigEarthNet-19} and \textit{BigEarthNet-43}).} For large-scale datasets, performance is primarily driven by architecture choice. Despite abundant training data, the ability to learn meaningful visual patterns remains crucial. SwinT (transformer-based model) achieves the best results, while modern convolutional architectures such as DenseNet and ResNet also perform consistently well (Figure~\ref{fig:exp2_marginal_arch_ben43}). Initialization plays a secondary role: although pretraining improves performance, the gains are limited and depend on the selected architecture (Figure~\ref{fig:exp2_marginal_init_ben43} and Figure~\ref{fig:exp2_marginal_arch_init_ben43}).

\begin{figure*}[!t]
    \centering
    \begin{subfigure}[t]{0.32\textwidth}
        \centering
        \includegraphics[width=\linewidth]{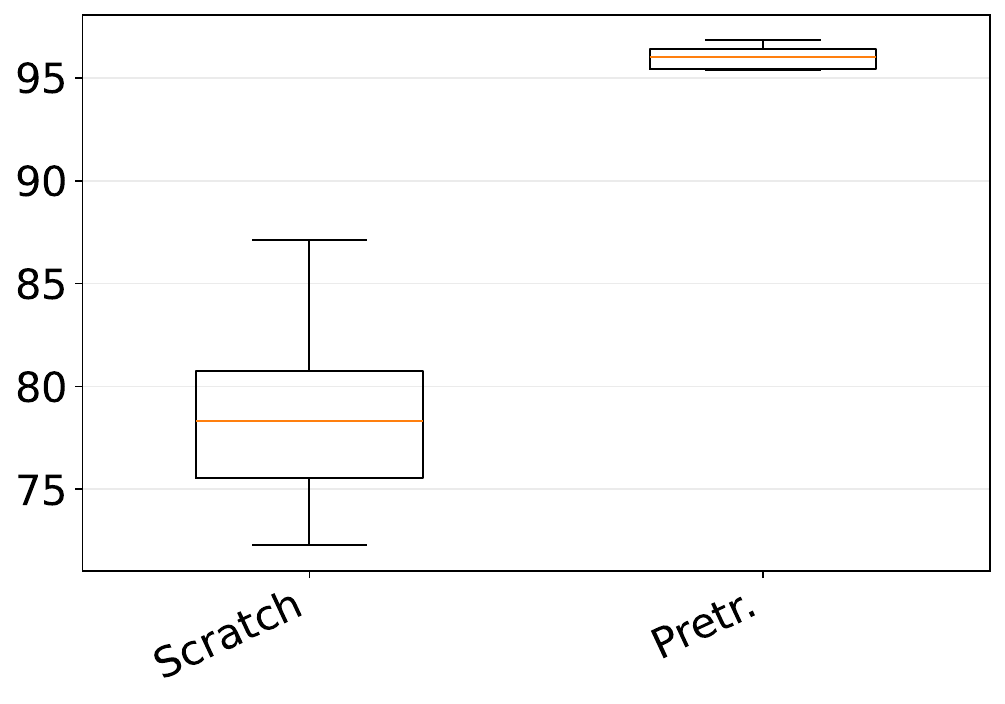}
        \caption{Initialization strategy}
        \label{fig:exp2_marginal_init_ucm}
    \end{subfigure}
    \begin{subfigure}[t]{0.32\textwidth}
        \centering
        \includegraphics[width=\linewidth]{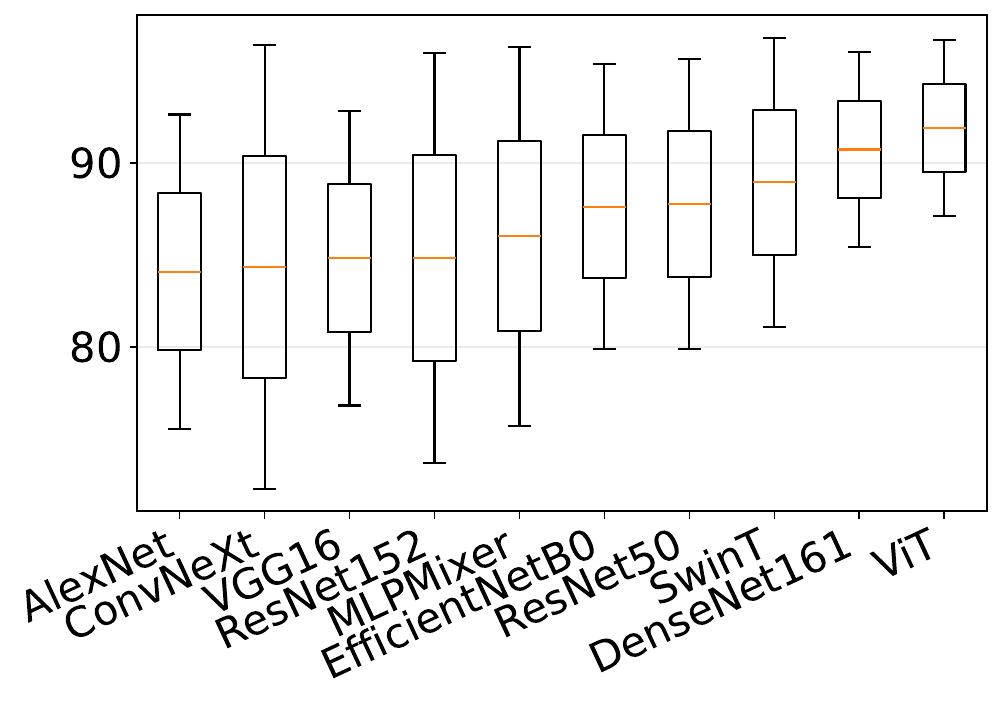}
        \caption{Architecture}
        \label{fig:exp2_marginal_arch_ucm}
    \end{subfigure}
        \begin{subfigure}[t]{0.32\textwidth}
        \centering
        \includegraphics[width=\linewidth]{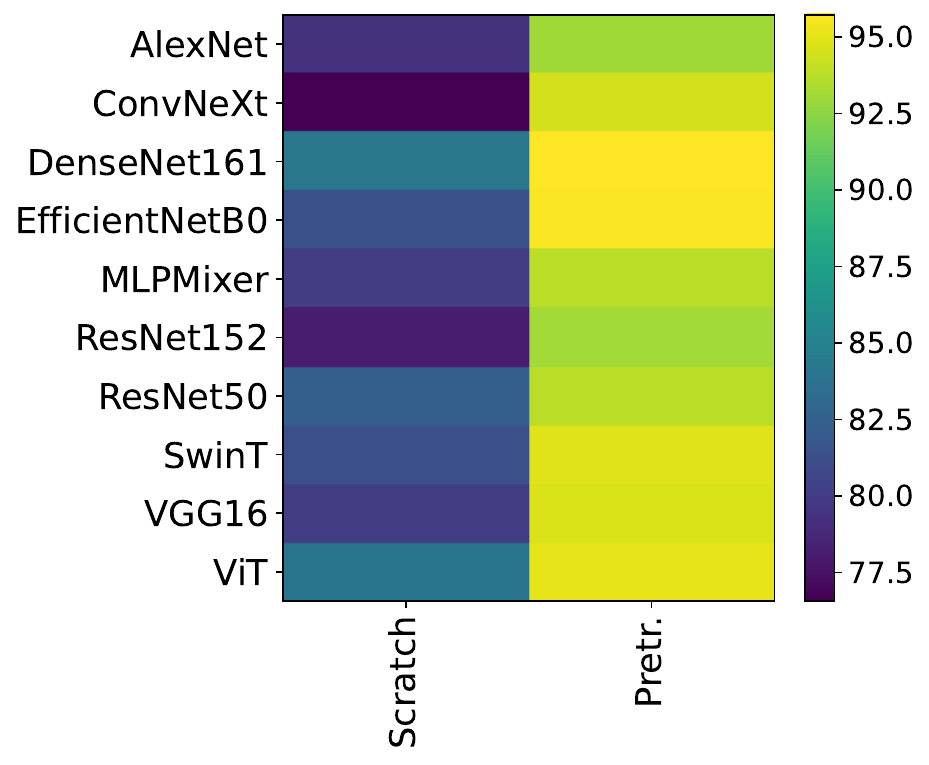}
        \caption{Architecture × Initialization}
        \label{fig:exp2_marginal_arch_init_ucm}
    \end{subfigure}

    \caption{Marginal effect of modules for \textit{UCM} (Cluster 2) in Scenario 2.}
    \label{fig:exp2_marginals_ucm}
\end{figure*}

\noindent\textbf{Cluster 2 (\textit{UCM} and \textit{AID}).}
In small datasets, performance is primarily driven by the initialization strategy. Limited training data makes learning strong representations from scratch difficult, making pretraining on large-scale datasets particularly beneficial for generalization and reducing overfitting (Figure\ref{fig:exp2_marginal_init_ucm}). Architecture plays a secondary role: scratch-trained models perform similarly poorly regardless of architecture, while pretraining improves performance uniformly across architectures (Figure~\ref{fig:exp2_marginal_arch_ucm}). Although transformer-based models such as ViT and SwinT achieve slightly better results, convolutional architectures such as DenseNet remain competitive.
 
\begin{figure*}[!t]
    \centering
    \begin{subfigure}[t]{0.32\textwidth}
        \centering
        \includegraphics[width=\linewidth]{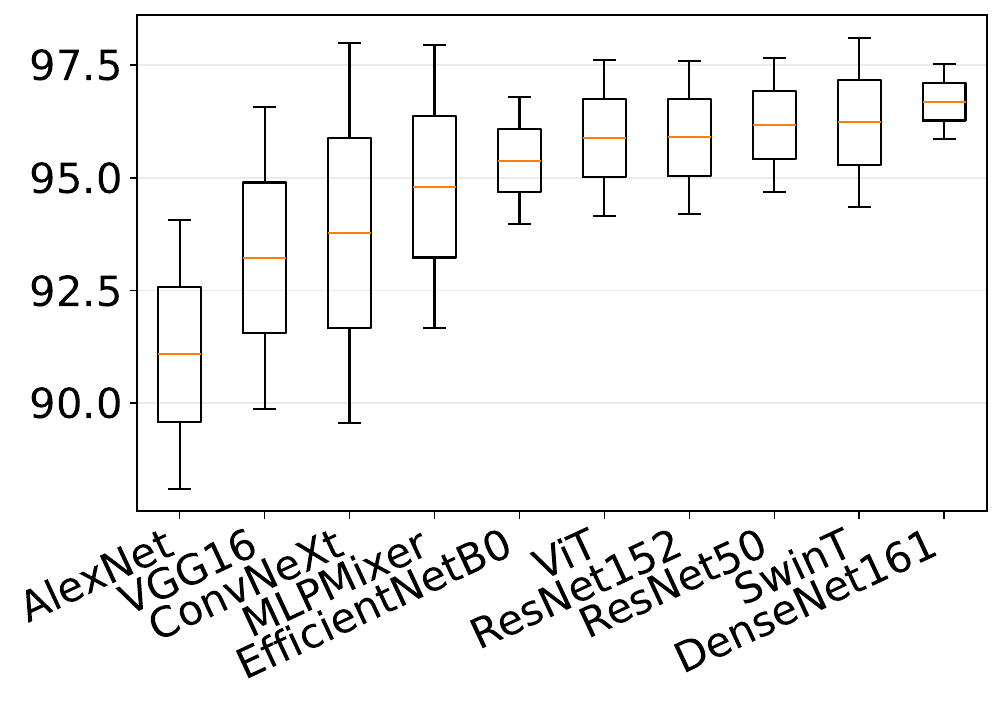}
        \caption{Architecture}
        \label{fig:exp2_marginal_arch_dfc}
    \end{subfigure}
    \hfill
    \begin{subfigure}[t]{0.32\textwidth}
        \centering
        \includegraphics[width=\linewidth]{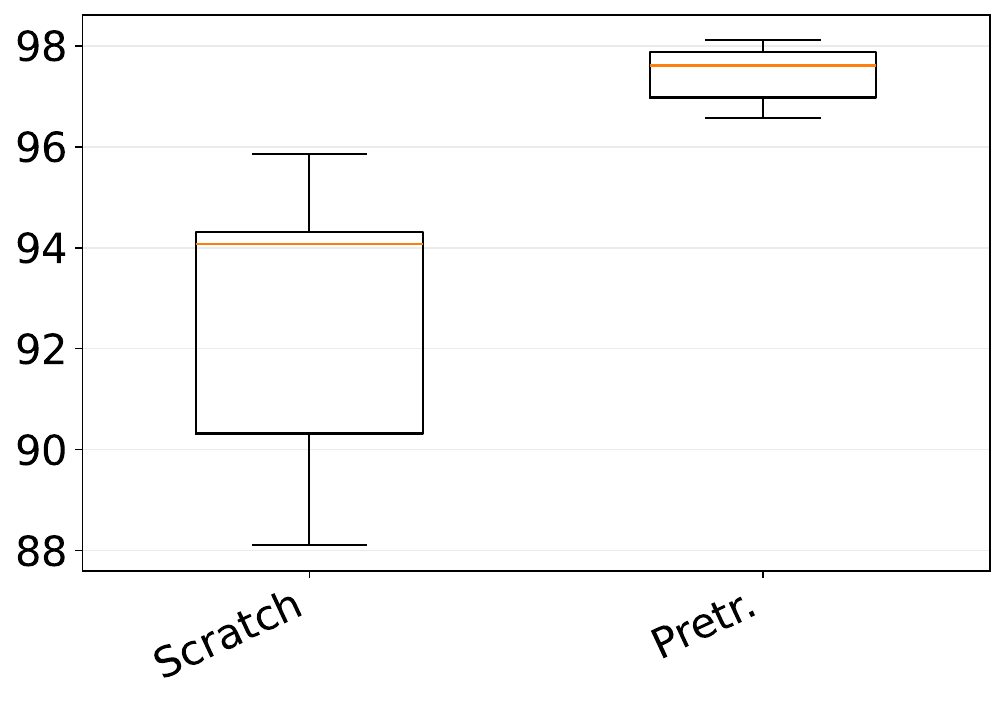}
        \caption{Initialization strategy}
        \label{fig:exp2_marginal_init_dfc}
    \end{subfigure}
    \hfill
    \begin{subfigure}[t]{0.32\textwidth}
        \centering
        \includegraphics[width=\linewidth]{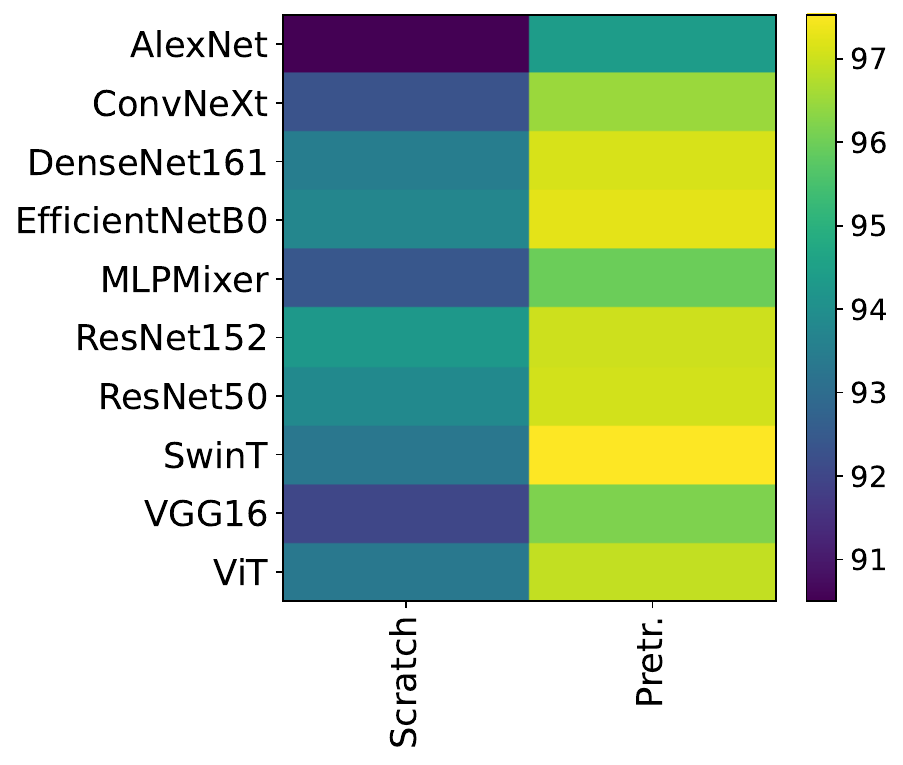}
        \caption{Architecture $\times$ Initialization}
        \label{fig:exp2_marginal_arch_init_dfc}
    \end{subfigure}

    \caption{Marginal effects of modules for \textit{DFC-15} (Cluster 3) in Scenario 2.}
    \label{fig:exp2_marginals_dfc}
\end{figure*}

\noindent\textbf{Cluster 3 (\textit{DFC-15}, \textit{PlanetUAS}, and \textit{MLRSNet}).} 
In this intermediate regime, both architecture and initialization strategy are important, and their interaction becomes non-negligible (Figure~\ref{fig:exp2_marginal_arch_init_dfc}): some architectures, such as ConvNeXt, VGG16, and MLPMixer, benefit more strongly from pretraining than others, such as ResNet152, indicating that a careful joint configuration of both factors is necessary.

\section{Extending Previous Benchmarking Knowledge}
The original benchmarking studies can be reinterpreted when considering dataset-specific sensitivities revealed by fANOVA. In Scenario~1, the original study reported a general preference for fine-tuning and the superiority of architectures as EfficientNet-B2. We show that these conclusions are dataset-dependent: fine-tuning dominates for large-scale datasets as \textit{BigEarthNet}, architectural choices and their interaction with the learning strategy become critical in data-limited regimes as \textit{Ankara}, and performance in medium-scale datasets as \textit{UCM} emerges from the interaction of multiple design choices. In Scenario~2, although pretrained transformer models were reported to perform strongly overall, our analysis reveals three sensitivity regimes: architectural capacity dominates for large datasets (e.g., \textit{BigEarthNet}), initialization strategy is decisive for small datasets (e.g., \textit{UCM} and \textit{AID}), and both factors interact in intermediate regimes (e.g., \textit{DFC-15}, \textit{PlanetUAS}, and \textit{MLRSNet}). These findings show that benchmarking conclusions are not universal but depend on the characteristics of the dataset.

\section{Conclusion}
We introduced a framework for analyzing the sensitivity of deep learning performance to design choices in remote sensing multi-label classification. Using fANOVA in two benchmarking scenarios and seven datasets, we derived dataset meta-representations that capture how design choices and their interactions influence performance variability. The analysis reveals three regimes: large-scale datasets are primarily driven by fine-tuning and architectural capacity, medium-scale datasets depend on the interaction of multiple design choices, and data-limited regimes are governed by architecture and its interaction with the learning strategy. The regimes align with intrinsic dataset meta-features, indicating that dataset properties strongly shape design-choice sensitivity. fANOVA-based meta-representations provide dataset-aware insights that go beyond traditional benchmarking rankings. Future work will investigate predicting sensitivity profiles from dataset meta-features and extend the analysis to modern visual foundation models. \noindent\textbf{Limitations}: The analysis is limited to CNN and early transformer architectures, with small number of configurations (up to three modules), and excludes recent visual foundation models and remote sensing–specific pretrained models, which may exhibit different sensitivity patterns. Also, performance data are reused from prior studies~\cite{dimitrovski2023current,stoimchev2023deep}, so training elements (e.g., augmentation, optimizer) are fixed and cannot be analyzed as additional design choices. Finally, clustering is performed on seven datasets per scenario. More datasets would improve statistical robustness of the silhouette score and of correlation or regression analysis relating design-choice importance to meta-features.

\section*{Acknowledgment}
This work was supported by the Horizon Europe ERA Chair AutoLearn-SI (101187010) and the Slovenian Research Agency through programs P2-0098 and P2-0103, projects J2-70078 and GC-0001, and a young researcher grant PR-12897 to AN.

%
%
%
%

\end{document}